\documentclass[11pt,a4paper]{article}
 
\usepackage[hyperref]{acl2021}
\usepackage{times}
\usepackage{latexsym}
\usepackage{amsmath}
\usepackage{graphicx}
\usepackage{booktabs}
\usepackage{multirow}

\usepackage[margin=1in]{geometry}
\usepackage{array}   
\usepackage{booktabs}

\usepackage[utf8]{inputenc}
\usepackage[T1]{fontenc}
\usepackage{xcolor}

\usepackage{fancyvrb}

\usepackage{microtype}

\aclfinalcopy 

\usepackage{todonotes}

\title{LettuceDetect: A Hallucination Detection Framework for RAG Applications}

\author{Ádám Kovács$^1$, Gábor Recski$^{1,2}$ \\
  $^1$ KR Labs \\
  $^2$ TU Wien\\
  \texttt{lastname@krlabs.eu}\\
  \texttt{firstname.lastname@tuwien.ac.at}}

\date{}

\begin{document}

\maketitle

\begin{abstract}
Retrieval-Augmented Generation (RAG) systems remain vulnerable to hallucinated answers despite incorporating external knowledge sources. We present \texttt{LettuceDetect}, a framework that addresses two critical limitations in existing hallucination detection methods: (1) the context window constraints of traditional encoder-based methods, and (2) the computational inefficiency of LLM-based approaches. Building on ModernBERT's extended context capabilities (up to 8k tokens) and trained on the RAGTruth benchmark dataset, our approach outperforms all previous encoder-based models and most prompt-based models, while being approximately 30 times smaller than the best models. \texttt{LettuceDetect} is a token-classification model that processes context-question-answer triples, allowing for the identification of unsupported claims at the token level. Evaluations on the RAGTruth corpus demonstrate an F1 score of 79.22\% for example-level detection, which is a 14.8\% improvement over Luna, the previous state-of-the-art encoder-based architecture. Additionally, the system can process 30 to 60 examples per second on a single GPU, making it more practical for real-world RAG applications.
 
\end{abstract}

\section{Introduction}

Large Language Models (LLMs) have made significant progress in recent years in terms of their performance \cite{Openai:2024, Grattafiori:2024, Gemma:2024}. However, the biggest obstacle to their usage in real-world applications is their tendency to hallucinate \cite{Kaddour:2023, Huang:2025}. Retrieval-Augmented Generation (RAG) is a method that enhances LLMs by supporting answers with context documents and retrieving knowledge from external sources, prompting the LLMs to ground their responses based on this information \cite{Gao:2024}. This technique is widely used to minimize hallucinations of LLMs. Despite the incorporation of context documents in RAG, LLMs continue to experience hallucinations \cite{Niu:2024}. 

Hallucinations are defined as outputs that are nonsensical, factually incorrect, or inconsistent with the provided evidence \cite{Ji:2023}. \citet{Ji:2023} categorizes these errors into two types: \textbf{Intrinsic hallucinations}, which arise from the model's inherent knowledge, and \textbf{Extrinsic hallucinations}, which occur when responses fail to be grounded in the provided context, such as in the case of RAG hallucinations \cite{Niu:2024}. While RAG can mitigate intrinsic hallucinations by grounding LLMs in external knowledge, extrinsic hallucinations persist due to imperfect retrieval processes or the model's tendency to prioritize its intrinsic knowledge over external context \cite{Sun:2025}, leading to factual contradictions. As LLMs remain prone to hallucinations, their utilization in high-risk settings, such as medical or legal fields, may be jeopardized \cite{Lozano:2023, Magesh:2024}.

\begin{figure}[ht]
    \centering
    \includegraphics[width=\linewidth]{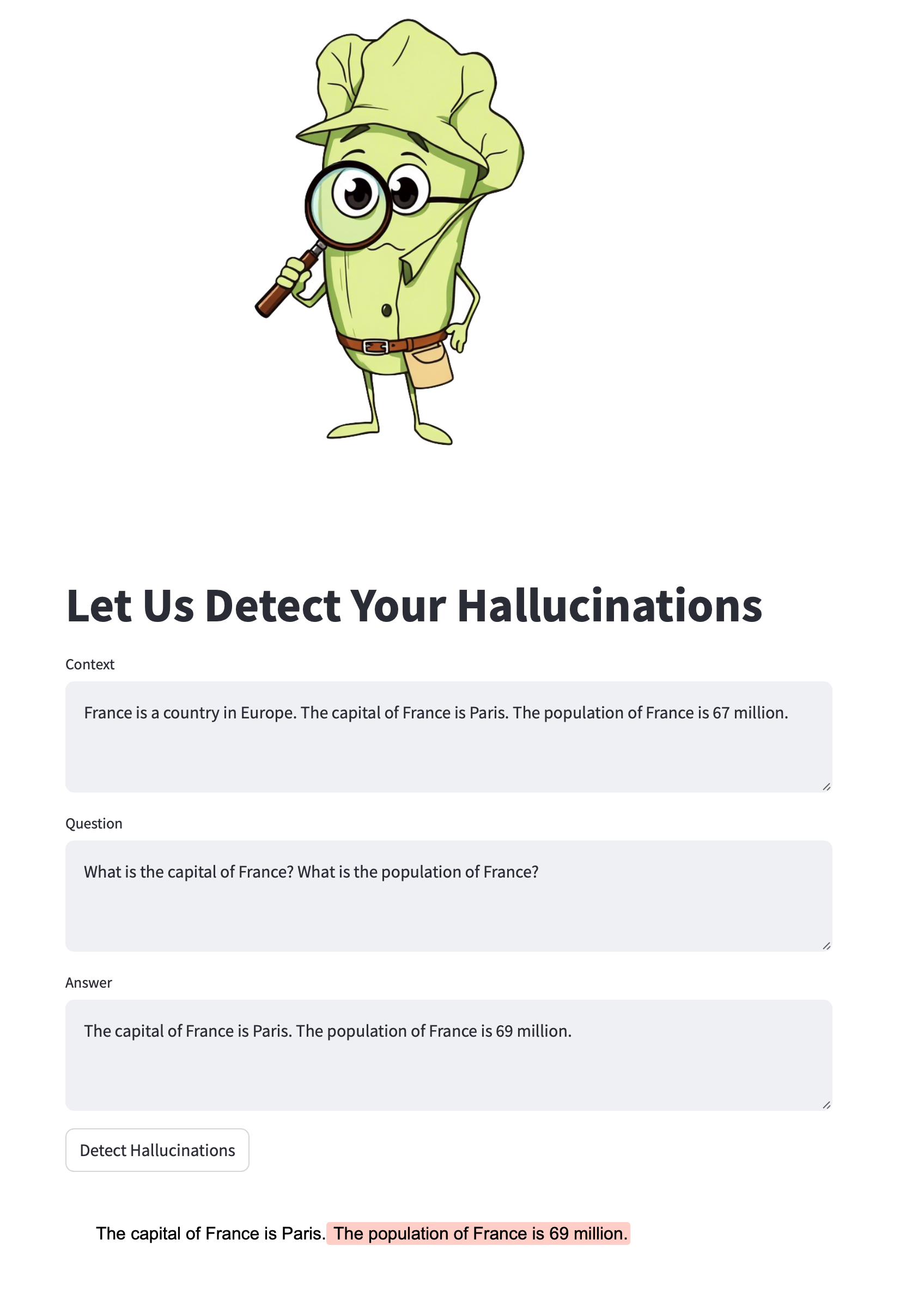}
    \caption{A web demo of our application built in Streamlit\footnote{\url{streamlit.io}}. It features three input fields: question, context, and answer. The output shows the highlighted hallucinated spans.}
    \label{fig:demo}
\end{figure}

We present \texttt{LettuceDetect}, a hallucination detection framework that utilizes ModernBERT \cite{Warner:2024}. Our approach trains a token-classification model to predict whether a token is supported by context documents and a question, determining if it is hallucinated. We frame this task as predicting tokens in the answers generated by large language models (LLMs), based on the provided context documents and the posed question. Our models are trained using the RAGTruth dataset \cite{Niu:2024}. The architecture we employ is similar to Luna \cite{Belyi:2025}, as we train an encoder-based model for this task. A demonstration of our web application is displayed in Figure~\ref{fig:demo}.

All components of our system are released under an MIT license and can be accessed on GitHub\footnote{\url{https://github.com/KRLabsOrg/LettuceDetect}} and via \texttt{pip} by installing the \texttt{lettucedetect}\footnote{\url{https://pypi.org/project/lettucedetect/}} package.

The trained models are published on Hugging Face also under MIT licenses. We have made available both a large model \footnote{\url{https://huggingface.co/KRLabsOrg/lettucedect-large-modernbert-en-v1}} and a base model \footnote{\url{https://huggingface.co/KRLabsOrg/lettucedect-base-modernbert-en-v1}}.

We believe our contribution will be valuable to the community, particularly since many effective hallucination detection methods are either under non-permissive licenses or depend on larger LLM-based models.

The remainder of this paper is structured as follows: Section~\ref{sec:rel} reviews recent methods for hallucination detection. Section~\ref{sec:method} details our method for training an encoder-based hallucination detection model built on ModernBERT. Section~\ref{sec:results} presents our findings on the example and span-level tasks using the RAGTruth dataset.

\section{Related work}
\label{sec:rel}

\paragraph{ModernBERT}

BERT \cite{Devlin:2019} was one of the first major successes of applying the Transformer
architecture \cite{Vaswani:2017} to natural language understanding. BERT uses only the Transformer's
encoder blocks in a bidirectional fashion, allowing it to learn context from both directions. As a result, BERT quickly
became the backbone of many NLP pipelines for tasks like classification, question answering, named entity recognition, etc.

BERT's initial design included certain limitations, such as a maximum sequence length of 512 tokens and less efficient attention mechanisms, leaving
room for architectural upgrades and larger-scale training. Despite the current rise of popularity of LLM-based architectures in NLP, such as GPT-4 \cite{Openai:2024}, Mistral \cite{Jiang:2023} or Llama-3 \cite{Grattafiori:2024}, encoder-based models are still widely used in many applications, because of their much smaller size and better-suited inference requirements that make them suitable for real-world applications.

ModernBERT \cite{Warner:2024} is a state-of-the-art encoder-only transformer architecture that incorporates several
modern design improvements over the original BERT model. It utilizes several enhancements, including rotary positional embeddings (RoPE) \cite{Su:2024} instead of traditional absolute positional embeddings. Additionally, it features an alternating local-global attention mechanism as described in \cite{Gemma:2024}, allowing it to efficiently manage sequences of up to 8,192 tokens. This makes it significantly more effective for long-context tasks, such as modern information retrieval \cite{Nussbaum:2025, Zhang:2024}. ModernBERT features a hardware-aware design and an expanded training corpus of 2 trillion tokens, including textual and code data. As a result, it achieves superior performance on various downstream benchmarks, such as GLUE for classification and BEIR for retrieval (while also maintaining faster inference speed) \cite{Nussbaum:2025, Zhang:2024}. Based on these findings, the main part of our paper is to use the advancements of ModernBERT in the hallucination detection of LLMs in an RAG setting. In this domain, long-context awareness is an inevitable feature.

\paragraph{Hallucination Detection} can vary in granularity, ranging from example-based detection (which assesses if an answer contains hallucinations) to token, span, or sentence-level detection \cite{Niu:2024}. The methods for detecting hallucinations also differ based on the techniques employed.

\paragraph{Prompt-based Techniques} typically utilize zero or few-shot large language models (LLMs) to identify hallucinations in LLM-generated responses. Few-shot or fine-tuned evaluation frameworks, such as RAGAS \cite{Es:2024}, Trulens\footnote{\url{https://www.trulens.org/}}, and ARES \cite{SaadFalcon:2024}, have emerged to provide hallucination detection at scale using LLM judges. However, real-time prediction remains a challenge for these methods. Other prompt-based approaches, like the zero-shot method SelfCheckGPT \cite{Manakul:2023}, employ stochastic sampling to identify inconsistencies across multiple response variants. Rather than relying on a single prompt, Chainpoll \cite{Friel:2023} implements a series of verification steps to detect hallucinations. \citet{Cohen:2023} presents a method of cross-examination between two LLMs to uncover inconsistencies. \citet{Chang:2024} utilized LLM-based classifiers trained on synthetic errors to detect both hallucinations and coverage errors in LLM-generated responses.

\paragraph{Fine-tuned LLM Judges} approaches involve training LLMs on hallucination detection tasks using specific training data. \citet{Niu:2024} not only introduced the RagTruth data but also presented a fine-tuned \texttt{Llama-2-13B} LLM, which achieved state-of-the-art performance on their test set, even surpassing larger models like GPT-4. RAG-HAT \cite{Song:2024} introduced a novel approach called Hallucination Aware Tuning (HAT), which involves training models to generate detection labels and provide detailed descriptions of identified hallucinations. They created a preference dataset to facilitate Direct Preference Optimization (DPO) training. Fine-tuning through DPO results in SOTA performance on the RAGTruth test set.

\paragraph{Encoder-based Solutions} focus on addressing computational efficiency constraints through domain-specific adaptations. RAGHalu \cite{Zimmerman:2024} employs a two-tiered encoder model that utilizes binary classification at each layer, fine-tuning a Natural Language Inference (NLI) model based on DeBERTa \cite{He:2021}. The approach most similar to our work is Luna \cite{Belyi:2025}, which also builds on DeBERTa and NLI to create a lightweight long-context hallucination detection system capable of managing longer contexts effectively. Luna draws connections between detecting entailment in NLI tasks and identifying hallucinations. They fine-tuned on a large, cross-domain corpus of question-answering-based RAG samples, with annotations provided by GPT-4. During the inference phase, Luna conducts sentence- or token-level checks on each model's response against the retrieved passages, effectively flagging unsupported fragments. FACTOID \cite{Rawte:2024} introduces a Factual Entailment (FE) framework, which represents a new form of textual entailment aimed at locating hallucinations at the token or span level. Other approaches, such as ReDeEp \cite{Sun:2025}, introduce techniques to analyze internal model states for hallucination detection.

\section{Data}
\label{sec:data}
We trained and evaluated our models using the RAGTruth dataset \cite{Niu:2024}. RAGTruth is the first large-scale benchmark for evaluating hallucinations in RAG settings. The dataset contains 18,000 annotated examples at the span level across three tasks: question answering, data-to-text generation, and news summarization.

For the question answering task, data was sampled from the MS MARCO dataset \cite{Bajaj:2018}, where each question had up to three corresponding contexts. The authors then prompted LLMs to generate answers based on the retrieved passages. In the data-to-text generation task, LLMs were asked to generate reviews for sampled businesses from the Yelp Open Dataset \cite{Yelp:2021}. For the news summarization task, randomly selected documents were taken from the training set of the CNN/Daily Mail dataset \cite{See:2017}, and LLMs were prompted to create summaries.

For response generation, various LLMs were employed, including GPT-4-0613 \cite{Openai:2024}, Mistral-7B-Instruct \cite{Jiang:2023}, and selections from the Llama models, such as Llama-2-7b-chat and Llama-2-13B-chat \cite{Grattafiori:2024}. Each sample in the dataset includes one response from each model, resulting in six responses per sample in RAGTruth. 

The entire dataset was annotated by human evaluators, who marked annotations in the responses and provided rationales. RAGTruth categorizes hallucinations into types such as Evident Conflict, Subtle Conflict, Evident Introduction of Baseless Information, and Subtle Introduction of Baseless Information. For our training, we focused solely on the binary task of recognizing hallucinations without considering these specific categories. An example from the dataset can be seen in Table~\ref{tab:ragtruth}.

We tokenized RAGTruth and found that the mean token length is 801 tokens, with a median of 741 tokens, a minimum of 194, and a maximum of 2632 tokens. This highlights the need for long-context models, such as ModernBERT, to effectively detect hallucinations.

\begin{table}[ht]
\centering
\scriptsize  
\resizebox{\columnwidth}{!}{
\begin{tabular}{@{}p{2.2cm} p{5.8cm}@{}}
\toprule
\textbf{QUESTION} & \textbf{How to explain quartiles?} \\
\midrule
\textbf{REFERENCE} & \textbf{Passage 2:} 
Second quartile (Q2) which is more commonly known as median splits the data in half (50\%). Median divides the data into a lower half and an upper half. Third quartile (Q3), also known as upper quartile, \emph{splits lowest 75\% (or highest 25\%) of data} \\
& \dots \\[3pt]
\textbf{RESPONSE} & The first quartile (Q1) splits the lowest 25\% of the data, while the second quartile (Q2) splits the data into two equal halves, with the median being the middle value of the lower half. Finally, the third quartile (Q3) splits \emph{the highest 75\% of the data.} \\
& \dots \\[3pt]
\textbf{ANNOTATION} & 
\textbf{Span:} highest 75\% \\
& \textbf{Type:} Evident Conflict \\
& \textbf{Reason:} Original: “lowest 75\% or highest 25\%” \\
& Generative: “highest 75\%” \\
\bottomrule
\end{tabular}%
}
\caption{An example of RAGTruth data, including question, references, response, and annotations.}
\label{tab:ragtruth}
\end{table}

\section{Method} 
\label{sec:method}

\begin{figure}[ht]
      \includegraphics[width=0.49\textwidth]{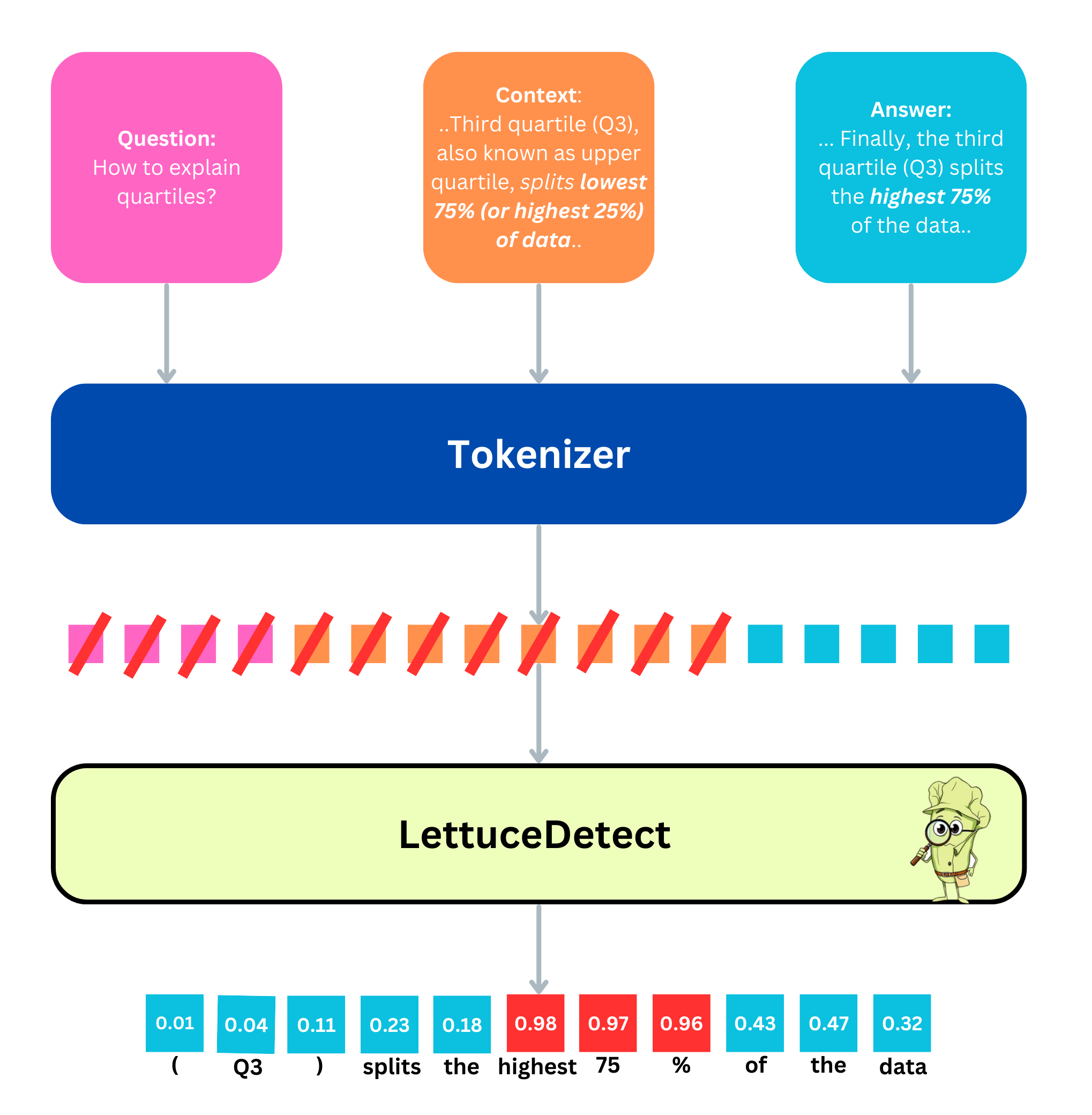}
  \caption{The architecture of LettuceDetect. The figure illustrates an example of a Question, Context, and Answer triplet as input to our architecture. After the tokenization step, the tokens are fed into LettuceDetect for token-level classification. Tokens from both the question and the context are masked (indicated by the red line) for loss calculations. In the output of LettuceDetect, we provide probabilities for each answer token. If the output type is span-level, we aggregate subsequent tokens that are hallucinated for the span-level output.}
    \label{fig:lettucedetect}
\end{figure}

We trained ModernBERT-base and -large variants as token classifiers on the RAGTruth dataset. Input sequences were constructed by concatenating context, question, and answer segments using special tokens ([CLS] for context, [SEP] for separation) and tokenized to a maximum length of 4,096 tokens (in the current version we haven't utilized ModernBERT's full 8,192 context length). For handling tokenization, we've used the \texttt{AutoTokenizer} \cite{Wolf:2020}. Our models are based solely on the ModernBERT architecture and were not pretrained on the NLI task, unlike previous encoder-based architectures.

The architecture leveraged Hugging Face's \texttt{AutoModelForTokenClassification} \cite{Wolf:2020} with ModernBERT as the backbone, and a classification head on top. Context/question tokens were masked (label=-100), while answer tokens were labeled as 0 (supported) or 1 (hallucinated). Training used AdamW optimization \cite{Loshchilov:2019} (learning rate $1\times10^{-5}$, weight decay 0.01) for 6 epochs on an NVIDIA A100 GPU. For data and batch handling, we've used PyTorch \texttt{DataLoader} \cite{Paszke:2019} (batch size=8, shuffling enabled). We evaluated models using token-level F1 score, saving the best-performing checkpoint via \texttt{safetensors}. Dynamic padding was implemented using \texttt{DataCollatorForTokenClassification} to process variable-length sequences efficiently. 

The final model predicts hallucination probabilities for each answer token, with span-level outputs generated by aggregating consecutive tokens exceeding a 0.5 confidence threshold. The best models are uploaded to huggingface. Our method can be seen in Figure~\ref{fig:lettucedetect}. We discuss the results in Section~\ref{sec:results}.

\begin{table*}[ht]
\centering
\scriptsize
\begin{tabular}{l|ccc|ccc|ccc|ccc}
\toprule
& \multicolumn{3}{c|}{\textbf{QUESTION ANSWERING}} 
& \multicolumn{3}{c|}{\textbf{DATA-TO-TEXT WRITING}} 
& \multicolumn{3}{c|}{\textbf{SUMMARIZATION}} 
& \multicolumn{3}{c}{\textbf{OVERALL}} \\
\cline{2-13}
\textbf{Method} 
& \textbf{Prec.} & \textbf{Rec.} & \textbf{F1} 
& \textbf{Prec.} & \textbf{Rec.} & \textbf{F1} 
& \textbf{Prec.} & \textbf{Rec.} & \textbf{F1} 
& \textbf{Prec.} & \textbf{Rec.} & \textbf{F1} \\
\midrule
Prompt\textsubscript{gpt-3.5-turbo}
& 18.8 & 84.4 & 30.8 
& 65.1 & 95.5 & 77.4 
& 23.4 & 89.2 & 37.1 
& 37.1 & 92.3 & 52.9 \\
Prompt\textsubscript{gpt-4-turbo}
& 33.2 & 90.6 & 45.6 
& 64.3 & 100.0 & 78.3 
& 31.5 & 97.6 & 47.6 
& 46.9 & 97.9 & 63.4 \\
SelCheckGPT\textsubscript{gpt-3.5-turbo}
& 35.0 & 58.0 & 43.7 
& 68.2 & 82.8 & 74.8 
& 31.1 & 56.5 & 40.1 
& 49.7 & 71.9 & 58.8 \\
LMvLM\textsubscript{gpt-4-turbo} 
& 18.7 & 76.9 & 30.1 
& 68.0 & 76.7 & 72.1 
& 23.2 & 81.9 & 36.2 
& 36.2 & 77.8 & 49.4 \\
Finetuned Llama-2-13B
& 61.6 & 76.3 & 68.2 
& 85.4 & 91.0 & 88.1 
& 64.0 & 54.9 & 59.1 
& 76.9 & 80.7 & 78.7 \\
RAG-HAT
& 76.5 & 73.1 & \textbf{74.8 }
& 92.9 & 90.3 & \textbf{91.6} 
& 77.7 & 59.8 & \textbf{67.6} 
& 87.3 & 80.8 & \textbf{83.9} \\
\midrule
ChainPoll\textsubscript{gpt-3.5-turbo}
& 33.5 & 51.3 & 40.5
& 84.6 & 35.1 & 49.6
& 45.8 & 48.0 & 46.9
& 54.8 & 40.6 & 46.7 \\
RAGAS Faithfulness
& 31.2 & 41.9 & 35.7
& 79.2 & 50.8 & 61.9
& 64.2 & 29.9 & 40.8
& 62.0 & 44.8 & 52.0 \\
Trulens Groundedness
& 22.8 & 92.5 & 36.6
& 66.9 & 96.5 & 79.0
& 40.2 & 50.0 & 44.5
& 46.5 & 85.8 & 60.4 \\
\textit{Luna} 
& 37.8 & 80.0 & 51.3 
& 64.9 & 91.2 & 75.9 
& 40.0 & 76.5 & 52.5 
& 52.7 & 86.1 & 65.4 \\
\midrule
\textbf{lettucedetect-base-v1}
& 60.64 & 71.25 & 65.52
& 89.30 & 86.53 & 87.89
& 53.89 & 47.55 & 50.52
& 76.64 & 75.50 & 76.07 \\
\textbf{lettucedetect-large-v1}
& 65.93 & 75.00 & \textbf{70.18}
& 90.45 & 86.70 & \textbf{88.54}
& 64.04 & 55.88 & \textbf{59.69}
& 80.44 & 78.05 & \textbf{79.22} \\
\bottomrule
\end{tabular}
\caption{Performance comparison at the example level across various tasks. We compare our results with models presented in Luna \cite{Belyi:2025} and RAGTruth \cite{Niu:2024}, as well as evaluation frameworks RAGAS and Trulens. The evaluation also includes a fine-tuned LLM from the RAG-HAT \cite{Song:2024} paper.}
\label{tab:results}
\end{table*}

\begin{table*}[ht]
\centering
\scriptsize
\begin{tabular}{l|ccc|ccc|ccc|ccc}
\toprule
& \multicolumn{3}{c|}{\textbf{QUESTION ANSWERING}} 
& \multicolumn{3}{c|}{\textbf{DATA-TO-TEXT WRITING}} 
& \multicolumn{3}{c|}{\textbf{SUMMARIZATION}} 
& \multicolumn{3}{c}{\textbf{OVERALL}} \\
\cline{2-13}
\textbf{Method}
& \textbf{Prec.} & \textbf{Rec.} & \textbf{F1}
& \textbf{Prec.} & \textbf{Rec.} & \textbf{F1}
& \textbf{Prec.} & \textbf{Rec.} & \textbf{F1}
& \textbf{Prec.} & \textbf{Rec.} & \textbf{F1} \\
\midrule
\textit{Prompt Baseline}\textsubscript{gpt-3.5-turbo}
& 7.9 & 25.1 & 12.1
& 8.7 & 45.1 & 14.6
& 6.1 & 33.7 & 10.3
& 7.8 & 35.3 & 12.8 \\
\textit{Prompt Baseline}\textsubscript{gpt-4-turbo}
& 23.7 & 52.0 & 32.6
& 17.9 & 66.4 & 28.2
& 14.7 & 65.4 & 24.3
& 18.4 & 60.9 & 28.3 \\
\textit{Finetuned Llama-2-13B}
& 55.8 & 60.8 & 58.2
& 56.5 & 50.7 & 53.5
& 52.4 & 30.8 & 38.6
& 55.6 & 50.2 & 52.7 \\
\midrule
\textbf{lettucedetect-base-v1}
& 62.65 & 60.40 & \textbf{61.50}
& 58.24 & 56.57 & \textbf{57.39}
& 52.98 & 28.08 & 36.71        
& 59.36 & 52.01 & \textbf{55.44}
\\
\textbf{lettucedetect-large-v1}
& 66.85 & 62.14 & \textbf{64.41}
& 64.71 & 55.99 & \textbf{60.04}
& 60.17 & 35.47 & \textbf{44.63}
& 64.92 & 53.96 & \textbf{58.93}
\\
\bottomrule
\end{tabular}

\caption{Performance comparison at the span level across different tasks. We compare our results with models presented in RAGTruth \cite{Niu:2024}. We limit this comparison to these papers, as other studies have not evaluated their performance on the span level task.}
\label{tab:span_results}
\end{table*}

\section{Evaluation}
\label{sec:results}

We evaluate our models using the RAGTruth test data across all task types, including question answering (QA), data-to-text, and summarization. Following the methodology outlined in \cite{Niu:2024}, we report both example-level and span-level detection performance, reporting precision, recall, and F1 score. Our models are compared against state-of-the-art baselines presented in \cite{Niu:2024, Song:2024, Belyi:2025}. This includes comparisons with prompt-based methods, such as gpt-4-turbo and gpt-3.5-turbo, as well as fine-tuned LLMs that have shown state-of-the-art performance on the RAGTruth data, including the previously established state-of-the-art model in \cite{Niu:2024} (a fine-tuned Llama-2-13B) and the current best result from \cite{Song:2024} (a fine-tuned LLM based on Llama-3-8B trained through DPO training). We also compare our models with encoder-based approaches, similar to ours, including the token classifier method presented in \cite{Belyi:2025}, which is based on DeBERTa.

Table~\ref{tab:results} illustrates our results on the example-level task. Our large model (lettucedetect-large-v1) outperforms all prompt-based methods (gpt-4-turbo achieved an overall F1 score of 63.4\% compared to lettucedetect-large-v1's 79.22\%). It also surpasses the previous state-of-the-art encoder-based model, Luna (\textbf{65.4\% }vs. \textbf{79.22\%}), and the previously established state-of-the-art fine-tuned LLM presented in \cite{Niu:2024} (fine-tuned Llama-2-13B with 78.7\% vs. 79.22\%). The only model that exceeds our large model's performance is the current state-of-the-art fine-tuned LLM based on Llama-3-8B presented in the RAG-HAT paper \cite{Song:2024} (\textbf{83.9\% }vs. \textbf{79.22\%}). Our base model (lettucedetect-base-v1) also demonstrates strong performance across tasks while being half the size of the large model. Considering our model's compact size (150M for the base model and 396M for the large model) and its optimized architecture based on ModernBERT, it is capable of processing approximately 30 to 60 examples per second on a single GPU. Given this optimized inference speed, it only falls short compared to one larger model (8B Llama). Overall, our models are highly efficient while being about 30 times smaller in size.

In Table~\ref{tab:span_results}, we present our results on the span-level task. In this task, we evaluate the overlap between the gold spans and the predicted spans. Following the RAGTruth paper, we measured character-level overlap and calculated precision, recall, and F1 score. Our models achieved state-of-the-art performance, with the Llama-2-13B model reaching an overall F1 score of 52.7\%, while our large model achieved 58.93\% F1 score. Please note that we were unable to compare our results with RAG-HAT on this task because they did not measure at this level. Additionally, RAGTruth did not include this evaluation in their published code, so we relied on our own implementation for this analysis.

\section{Conclusion}
We present \texttt{LettuceDetect}, a lightweight and efficient framework for hallucination detection in RAG systems. By leveraging ModernBERT’s long-context capabilities, our baseline models achieve strong performance on the RAGTruth benchmark while remaining highly efficient in inference settings. This work serves as a foundation for our future research, where we plan to expand the framework to include more datasets, additional languages, and enhanced architectures. Even in its current form, LettuceDetect demonstrates that effective hallucination detection can be achieved with lean, purpose-built models.


\clearpage

\bibliographystyle{acl_natbib}
\bibliography{lettucedetect}


\end{document}